\documentclass[letterpaper, 10 pt, conference]{ieeeconf}  

\IEEEoverridecommandlockouts                              

\usepackage{graphics} 
\usepackage{epsfig} 
\usepackage{amsmath} 
\usepackage{amssymb}  

\usepackage{graphicx}
\usepackage{color}

\usepackage{tikz}
\usetikzlibrary{shapes,arrows}

\newtheorem{prop}{Proposition}
\newtheorem{definition}{Definition}
\newtheorem{remark}{Remark}

\title{\LARGE \bf Robot kinematic structure classification from time series of visual data}

\author{Alberto Dalla Libera$^{1}$, Matteo Terzi$^{2}$, Rossi Alessandro$^{1}$, Gian Antonio Susto$^{1}$ and Ruggero Carli$^{1}$
	\thanks{$^{1}$A. Dalla Libera, R. Carli, A. Rossi and G. A. Susto are with the Deptartment of Information Engineering, University of Padova, Via Gradenigo 6/B, 35131 Padova, Italy {\texttt\small [dallaliber@dei.unipd.it, carlirug@dei.unipd.it, rossiale@dei.unipd.it, gianantonio.susto@dei.unipd.it]}. $^{2}$M. Terzi is with the Human Inspired Technology Center, 35131 Padova, Italy {\texttt\small [matteo.terzi@phd.unipd.it]}.	
	}%
}

\begin{document}

\maketitle
\thispagestyle{empty}
\pagestyle{empty}

\begin{abstract}
In this paper we present a novel algorithm to solve the robot kinematic structure identification problem. Given a time series of data, typically obtained processing a set of visual observations, the proposed approach identifies the ordered sequence of links associated to the kinematic chain, the joint type interconnecting each couple of consecutive links, and the input signal influencing the relative motion. Compared to the state of the art, the proposed algorithm has reduced computational costs, and is able to identify also the joints' type sequence. 
\end{abstract}

\section{INTRODUCTION}

In recent years, there has been an increasing interest toward Modular Robotics and reconfigurable and adaptable robots have started to be designed \cite{Mod_rob_1, Mod_rob_2, Mod_rob_3}. In particular, reconfigurability and modularity can be exploited to build robots with greater adaptability to several different environments, as well as robots able to accomplish different tasks, obtaining important cost and time reductions.

On the other hand, Modular Robotics introduces new challenging issues. One of these is the need to handle robots with variable kinematic structures, where this variability might result in a partially or complete lack of knowledge about the kinematic structure. It is worth stressing that the forward and inverse kinematics are fundamental in robotics applications; remarkable examples are motion planning \cite{latombe}, robot modeling and control \cite{siciliano}.

When dealing with standard robots, the prior knowledge about the robot geometry is extremely precise, since most of the times CAD models are available and direct measurements of the robot parameters are possible. Kinematic models relate joint input signals and robot configurations. Typically they are computed as a sequence of relative transformations between reference frames properly assigned \cite{siciliano}. Unavoidable inaccuracies in the geometrical parameters, wear and systematic errors in the measurements, make sometimes calibration procedures necessary \cite{kin_calibration_PSW}.

In the Modular Robotics context, where the uncertainty about the robot geometry might be particularly high, the development of algorithms able to estimate the kinematic structure starting from a time series of visual observations is crucial. 
The kinematic structure identification problem is defined at different levels, depending on the sensors used and the amount of prior information available. When there is no prior knowledge, firstly it is necessary to identify the rigid bodies composing the robot, and extract information about their poses. Secondly, starting from this piece of information, the robot kinematic is learned identifying the ordered sequence of links, the type of the joint connecting any pair of consecutive links and the corresponding input signal.

The first level is strictly related to the kind of sensors used in the data collection. For example, in \cite{kin_point_cloud} point cloud data have been considered. More precisely, the authors have proposed to identify the links by clustering points based on their relative distances, and by assuming that each cluster corresponds to one link. The task is much more complicated when observations come from a standard 2D camera. Indeed in this case the clustering phase should be preceded by the identification of features which are constant over the observations.

In this paper, we consider the setup introduced in \cite{kin_end_effector, Sturm_object} and \cite{Sturm_robot}, where a distinct fiducial marker is attached to each link. This assumption, that simplifies considerably the data acquisition, is particularly reasonable in the Modular Robotics context.  

The focus of this paper is on the second level of the kinematic structure reconstruction. Specifically, starting from a time series containing the marker poses and the joint signals, we propose an algorithm able to reconstruct the ordered markers sequence associated to the robot kinematic chain, together with the joint types connecting subsequent links and the corresponding joint input signals.

Similar problems have been treated in \cite{kin_end_effector} and \cite{Sturm_robot}. In \cite{kin_end_effector}, the authors have restricted the scope of their work to the case of only revolute joints, and the learning of the kinematic parameters is obtained by a gradient-based minimization procedure. Moreover, numerical results highlight how the convergence is guaranteed only when each link has a marker attached to.
In \cite{Sturm_robot} the authors have proposed the use of Gaussian Processes (GP) methodologies. 
However, the focus is limited to learn only the sequence of markers, and no joint type identification has been considered. Moreover, even if the authors have proposed a strategy to simplify the standard GP procedure, the algorithm they developed might be quite expensive from a computational point of view, in particular when dealing with manipulator with a relevant number of degrees of freedom.

The kinematic structure identification  algorithm  we introduce is based on checking the feasibility of three systems of equations, which are obtained starting from elementary kinematic relations between pairs of subsequent links and using information extracted from time series of visual data. More precisely, given a couple of markers attached to subsequent links and the corresponding joint input signal, a linear system of equations holds true if the three elements define a prismatic transformation, instead a linear and a non linear systems are satisfied if the transformation is revolute. 
In general, it is possible to exhibit sets of observations for which systems of equations hold true though the pair of markers and the joint signal considered are not in relation among them. However, by extensive Monte Carlo simulations
we show that this \emph{false-positive} fact is very unlikely to appear. To ensure that the feasibility of the introduced systems of equations is a necessary and sufficient condition to verify if two markers are attached to consecutive links, we need to apply our strategy with data obtained from \emph{fully informative} sets of observations; in the paper we exhibit a class of trajectories from which it is possible to properly select observation sets which are fully informative. \\
Compared to the state of the art, the proposed approach is less expensive as regards the computational costs, since it is based on the solution of linear and non linear systems of equations with low dimensionality. Moreover, differently from \cite{Sturm_robot}, the proposed algorithm fully reconstructs the kinematic structure, including the joint type sequence.

The paper is organized as follows. In the next Section we describe the considered setup, and in Section \ref{sec:couple_relation} we derive the three systems of equations used in the kinematic identification. The \emph{fully informative} trajectories are reported in Section \ref{sec:trj}, while the proposed algorithm and the Monte Carlo simulations are in Section \ref{sec:proposed_approach}.

\section{SETUP DESCRIPTION AND PROBLEM FORMULATION} \label{sec:setup}
In this section we formally describe the setup and problem considered in this work.

\begin{figure}
	\centering
	\includegraphics[width=0.8\linewidth]{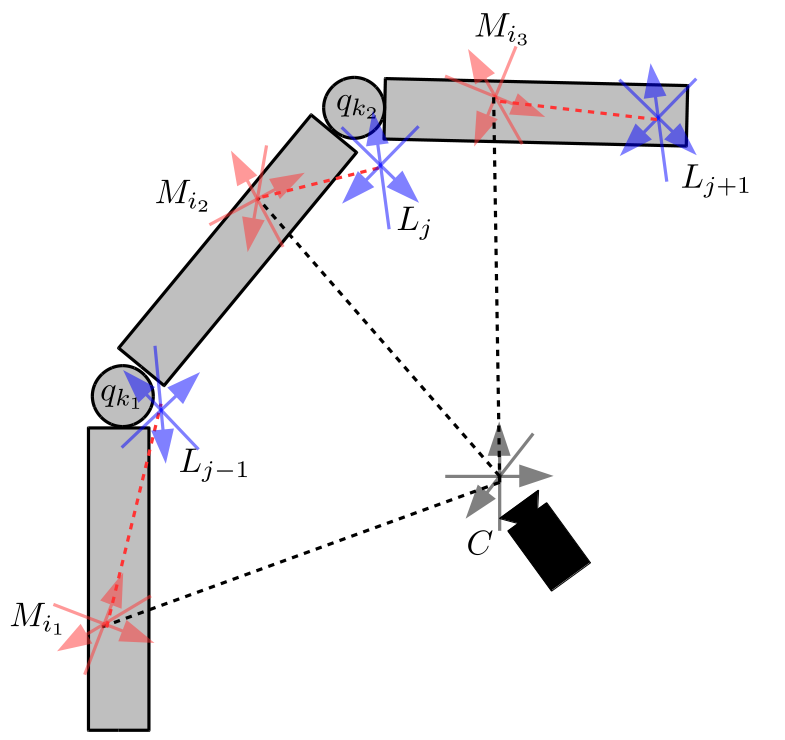}
	\caption{Symbolic representation of the setup.}
	\label{fig:setup}
\end{figure}

The framework is the same one adopted in \cite{Sturm_robot}, and it consists in a camera and a robotic arm, composed by $n$ links and $n-1$ joints, forming an open kinematic chain. As far as the camera placement is concerned, we adopted an eyes-to-hand configuration, namely the camera is observing the robot and its pose is fixed with respect to the world reference frame (RF). For a pictorial description of the overall setup see Figure \ref{fig:setup}.  

We assume that each robot link $L_j$, $j=1, \ldots, n$, has a fiducial marker $M_i$, $i=1, \ldots, n$, attached to it, see for instance, \cite{fiducial_markers, apriltag1} and \cite{apriltag2}. The $j$ subscript denotes the link position in the kinematic chain, in particular $L_1$ and $L_n$ correspond, respectively, to the base link and the last link.

The marker-link relations are unknowns and, for this reason we use different subscripts to enumerate markers and links. More formally, referring to the notation above introduced, given the $i$-th marker $M_i$, the link $L_j$ at which $M_i$ is attached to is unknown.

The relative poses between the world RF and each marker are obtained processing each single frame coming from the camera \cite{fiducial_markers, apriltag1, apriltag2}. Specifically, let $\boldsymbol{p_{M_i}^c}(t)$ be the pose of marker $M_i$ at time $t$, denoted with the respect to the camera RF (hereafter indicated by the superscript $c$); then
\begin{equation*}
\boldsymbol{p_{M_i}^c}(t) =
\begin{bmatrix}
\boldsymbol{l_{M_i}^c}(t)^T &
\boldsymbol{o_{M_i}^c}(t)^T
\end{bmatrix}^T 
\text{,}
\end{equation*}%
where $\boldsymbol{l_{M_i}^c}$ denotes the position vector, and it is composed by the three Cartesian coordinates $x_{M_i}^c$, $y_{M_i}^c$ and $z_{M_i}^c$, while, as far as the orientation $\boldsymbol{o_{M_i}^c}$ is concerned, the yaw-pitch-roll convention is adopted, where $\gamma_{M_i}^c(t)$, $\beta_{M_i}^c(t)$ and $\alpha_{M_i}^c(t)$ are respectively the yaw, pitch and roll angles. For computational reasons, it is convenient to express the relative orientation between $M_i$ and the camera using the rotation matrix $R_{M_i}^{c}$, that is related to the yaw, pitch and roll angles by the standard expression $R_{M_i}^c = R_z(\gamma_{M_i}^c)R_y(\beta_{M_i}^c)R_x(\alpha_{M_i}^c)$.
Similar definitions hold for $\boldsymbol{p_{L_j}^c}$, which denotes the pose of the $j$-th link.

Then, by processing the frame taken by the camera at time instant $t$, it is possible to reconstruct the set of poses $D_p(t)=\{\boldsymbol{p_{M_1}^c}(t), \cdots, \boldsymbol{p_{M_n}^c}(t)  \}$. 

In addition, we assume that the vector of the joints configurations $\boldsymbol{q}(t)= \left[q_{1}(t),\, \dots,\, q_{n-1}(t)\right]$ is available at time $t$; specifically, $q_{k}(t)$ parametrizes the relative displacement between the two links connected by the $k$-th joint. However, we assume that also the relations joint-links are unknown, namely, for the $k$-th joint we do not know which pair of links $L_{j_1}$ and $L_{j_2}$ is connected by joint $k$. For this reason, similarly to what done when considering the markers, we use different subscripts to denote joints and links.  

The main goal of this paper is that of identifying the robot kinematic structure starting from a time series of measurements $X = \{ (D_p(t_1),\boldsymbol{q}(t_1)),\dots, (D_p(t_T),\boldsymbol{q}(t_T)) \}$, composed by the joints configurations and the marker poses.

From now on, to keep the notation compact, we point out explicitly the dependencies on time only when it is necessary. 

The identification of the robot kinematic structure can be decomposed into three subtasks:
\begin{itemize}
	\item Identifying $S_M=\{M_{i_1},\dots,M_{i_n}\}$, i.e., the sequence of markers associated to the kinematic chain $L_1,\dots,L_n$; 
	\item Identifying $S_Q=\{Q_{k_1},\dots,Q_{k_{n-1}}\}$, i.e., the sequence of  joint types 
	connecting consecutive links along the kinematic chain, starting from the pair $(L_1,L_2)$, up to the pair $(L_{n-1},L_{n})$; more precisely $Q_{k_j}$ is a binary variable assuming value $0$ (resp. value $1$) when the joint connecting $L_j$ and $L_{j+1}$ is prismatic (resp. revolute);
	\item Identifying $S_q=\{q_{k_1},\dots,q_{k_{n-1}}\}$, i.e., the sequence of joint signals that parametrize the corresponding transformations in $S_Q$. 
\end{itemize}

\section{RELATIONS BETWEEN COUPLE OF SUBSEQUENT MARKERS}\label{sec:couple_relation}
In this section we provide useful expressions that describe the relative motion between markers which are attached to consecutive links; in particular we will distinguish the case where the joint connecting the links is prismatic from the case where it is revolute.

More formally, let $M_{i_1}$ and $M_{i_2}$ be the two markers, and let $L_{j_1}$ and $L_{j_2}$ be the links they are attached to. Supposing $j_1=j_2-1$ (similar considerations hold for $j_1=j_2+1$), we will assign a RF to each marker and to each link and, based on these RFs, we will provide a mathematical description of the relative motion between $M_{i_1}$ and $M_{i_2}$. In particular, given a set of observations $X$, we will identify three systems of equations, that will be exploited in the next sections, to discriminate if the links associated to  $M_{i_1}$ and $M_{i_2}$ are directly connected or not. Additionally, if  $L_{j_1}$ and $L_{j_2}$ are connected, the systems will uniquely identify the joint type connecting the two links and also the joint variable $q_k$ describing the relative displacement. 
 



\subsection{Reference frames conventions}

The definition of a RF for each link and for each maker is required to provide a mathematical description of the transformations occurring along the kinematic chain. As far as the links are concerned, we adopt the Denavit-Hartenberg (DH) convention; for details we refer the interested reader to \cite{siciliano}, chapter $2.8.2$. Once the RFs of the links have been assigned, the expression of $R_{L_{j-1}}^{L_{j}}$, i.e., the relative orientation between the consecutive links $L_{j-1}$ and $L_j$, is given by
\begin{equation}
R_{L_{j}}^{L_{j-1}} = R_z(\theta_j)R_x(\alpha_j) \text{,} \label{eq:joint_orinentation}
\end{equation}%
being $R_z$ and $R_x$ the elementary rotation matrices around the $z$-axis and the $x$-axis, respectively, and $\alpha_j$ a constant parameter (see \cite{siciliano}). In case the joint connecting $L_{j-1}$ and $L_{j}$ is prismatic, then $\theta_j$ is constant and equal to $\theta_j^0$, while, if the joint is revolute and controlled by $q_k$, it holds $\theta_j = \theta_j^0 + q_k$. The relation between the relative positions of $L_{j-1}$ and $L_j$ is described by $\boldsymbol{l_{L_{j}}^{L_{j-1}}}$, i.e., the expression of the origin of $L_{j}$ with the respect to $L_{j-1}$, and is given by
\begin{equation}
\boldsymbol{l_{L_{j}}^{L_{j-1}}} =
\begin{bmatrix}
0\\0\\d_j
\end{bmatrix} +  R_z(\theta_j)\begin{bmatrix}
a_j\\0\\0
\end{bmatrix} \text{,} \label{eq:joint_traslation}
\end{equation}%
where $R_z(\theta_j)$ is defined as before and $a_j$ is a constant parameter of the kinematic (see \cite{siciliano}). If the joint connecting $L_{j-1}$ and $L_{j}$ is revolute then $d_j$ is constant and equal to $d_j^0$, while, if it is prismatic and parametrized by $q_k$, then it holds $d_j=d_j^0+q_k$.

Additionally, we need to define position and orientation of the reference frame of each marker with the respect to the reference frame of the link they are attached to. For example, suppose that $M_{i_1}$ is attached to $L_{j_1}$, then position and orientation of $M_{i_1}$ w.r.t. $L_{j_1}$ are described, respectively, by $\boldsymbol{l_{M_{i_1}}^{L_{j_1}}}$ and $R_{M_{i_1}}^{L_{j_1}}$. For later use it is convenient to introduce also $R_{L_{j_1}}^{M_{i_1}} = (R_{M_{i_1}}^{L_{j_1}})^T$ and  $\boldsymbol{l_{L_{j_1}}^{M_{i_1}}}$. Similar definitions hold for $M_{i_2}$ and $L_{j_2}$.

Since the marker-link transformations are fixed, $\boldsymbol{l_{M_{i_1}}^{L_{j_1}}}$ and $R_{M_{i_1}}^{L_{j_1}}$ are independent from the joint values $\boldsymbol{q}$ and constant over the time. 

It is worth stressing that $\boldsymbol{l_{M_{i_1}}^{L_{j_1}}}$ and $R_{M_{i_1}}^{L_{j_1}}$ are unknown, and we did not introduce any limitation on the way the markers are attached to links. From a practical point of view, this fact is very interesting, since it allows to adopt the proposed algorithm even in setups different from the one we described in Section \ref{sec:setup}. Indeed, it might happen that the markers are not available and the use of ad-hoc computer-vision algorithms is required to get information about the robot displacements \cite{marker_from_feature_1, marker_from_feature_2, marker_from_feature_3}. In this context the markers placement is not controllable, but it still holds that $\boldsymbol{l_{M_{i_1}}^{L_{j_1}}}$ and $R_{M_{i_1}}^{L_{j_1}}$ are constant. 

Let $\boldsymbol{l_{M_{i_2}}^{M_{i_1}}}$ and $R_{M_{i_2}}^{M_{i_1}}$ be, respectively, the Cartesian coordinates of the origin of $M_{i_2}$ w.r.t. $M_{i_1}$, and the relative orientation between $M_{i_2}$ and $M_{i_1}$. Assuming that $L_{j_1}$ and $L_{j_2}$ are subsequent, i.e., $j_2=j_1+1$, $\boldsymbol{l_{M_{i_2}}^{M_{i_1}}}$ is given by
\begin{equation}
\boldsymbol{l_{M_{i_2}}^{M_{i_1}}} = \boldsymbol{l_{L_{j_1}}^{M_{i_1}}} +R_{L_{j_1}}^{M_{i_1}}\boldsymbol{l_{L_{j_2}}^{L_{j_1}}} + R_{M_{i_2}}^{M_{i_1}}\left( -\boldsymbol{l_{L_{j_2}}^{M_{i_2}}} \right) \text{.} \label{eq:relative_position_consecutive markers}
\end{equation}%
Moreover, exploiting standard properties of rotation matrices, the following equation holds
\begin{equation}
R_{M_{i_2}}^{M_{i_1}} = R_{L_{j_1}}^{M_{i_1}} R_{L_{j_2}}^{L_{j_1}}(\theta_{j_2}) R_{M_{i_2}}^{L_{j_2}}\label{eq:relative_orientation_consecutive markers} \text{,}
\end{equation}%
It is worth remarking that, in the described setup, we have knowledge of $\boldsymbol{l_{M_{i_2}}^{M_{i_1}}}$, since $\boldsymbol{l_{M_{i_2}}^{M_{i_1}}} = (R_{M_{i_1}}^{c})^T (\boldsymbol{l_{M_{i_2}}^{c}}-\boldsymbol{l_{M_{i_1}}^{c}})$, where $R_{M_{i_1}}^{c}$, $\boldsymbol{l_{M_{i_2}}^{c}}$ and $\boldsymbol{l_{M_{i_1}}^{c}}$ are obtained processing the information coming from the camera. Moreover, also $R_{M_{i_2}}^{M_{i_1}} = (R_{M_{i_1}}^{c})^T R_{M_{i_2}}^{c}$ is known, since by definition it is a function of the camera observations $R_{M_{i_1}}^{c}$ and $R_{M_{i_2}}^{c}$.

In the remaining part of this Section we further investigate the above relations, distinguishing the case where the joint connecting two successive links is prismatic from the case where the joint is revolute.

\subsection{Prismatic joint}
Assume that the joint connecting $L_{j_1}$ and $L_{j_2}$ is prismatic. Since in this case the angle $\theta_j$ is constant, it follows that the relative orientation between $L_{j_1}$ and $L_{j_2}$ is not affected by variations of the joint variable, that is, the matrix $R_{M_{i_2}}^{M_{i_1}}$ is also constant over the time.
In addition, assume that $q_k$ is the joint variable associated to the prismatic joint connecting $L_{j_1}$ and $L_{j_2}$. By substituting the expression of $\boldsymbol{l_{L_{j_2}}^{L_{j_1}}}$ given in \eqref{eq:joint_traslation} into \eqref{eq:relative_position_consecutive markers}, the following equation holds
\begin{align}
\boldsymbol{l_{M_{i_2}}^{M_{i_1}}} = & \boldsymbol{l_{L_{j_1}}^{M_{i_1}}} + R_{L_{j_1}}^{M_{i_1}} \left(\begin{bmatrix}
0\\0\\d^0_j + q_k
\end{bmatrix} +  R_z(\theta^0_j)\begin{bmatrix}
a^0_j\\0\\0
\end{bmatrix}\right) \notag \\
&+ R_{M_{i_2}}^{M_{i_1}}\left( -\boldsymbol{l_{L_{j_2}}^{M_{i_2}}} \right) \notag \\
=& \boldsymbol{l_{L_{j_1}}^{M_{i_1}}} + R_{L_{j_1}}^{M_{i_1}} \left(\begin{bmatrix}
0\\0\\d^0_j
\end{bmatrix} +  R_z(\theta^0_j)\begin{bmatrix}
a^0_j\\0\\0
\end{bmatrix}\right) \notag \\
&+ R_{M_{i_2}}^{M_{i_1}}\left( -\boldsymbol{l_{L_{j_2}}^{M_{i_2}}} \right) + R_{L_{j_1}}^{M_{i_1}}\begin{bmatrix}0\\0\\q_k\end{bmatrix} \text{,} \label{eq:non_lin_prims_position}
\end{align}%
where the first three terms of the last equation are constant and they can be compacted in the vector $\boldsymbol{l_{i_1,i_2}^{*}}$, while the last term depends on the joint coordinate $q_k$.

Observe that the last equation defines a system of equations which are linear w.r.t. $\boldsymbol{l_{i_1,i_2}^{*}}$ and the third column of $R_{L_{j_1}}^{M_{i_1}}$ that we denote hereafter as $Z\left( R_{L_{j_1}}^{M_{i_1}} \right)$. Specifically we can write
\begin{equation}\label{eq:linear}
\boldsymbol{l_{M_{i_2}}^{M_{i_1}}} = \begin{bmatrix}I_3  & q_k I_3\end{bmatrix} \begin{bmatrix}\boldsymbol{l_{i_1,i_2}^{*}}  \\ Z\left( R_{L_{j_1}}^{M_{i_1}} \right)\end{bmatrix} = A(q_k)\boldsymbol{b_{i_1,i_2}} \; \text{,}
\end{equation}%
where $A(q_k)= \left[ I_3  \,\, q_k I_3\right]$ and $\boldsymbol{b_{i_1,i_2}} = \left[ \left(\boldsymbol{l_{i_1,i_2}^{*}}\right)^T  \,\, Z\left( R_{L_{j_1}}^{M_{i_1}} \right)^T \right]^T$ subject to the constraint 
\begin{equation}
Z\left( R_{L_{j_1}}^{M_{i_1}} \right)^T Z\left( R_{L_{j_1}}^{M_{i_1}} \right)= 1 \text{.} \label{eq:prism_constr}
\end{equation}%
Note that Equation \eqref{eq:linear} holds when we have only one observation. In general, if the set of observations $X$ has cardinality $T$, Equation \eqref{eq:linear} is replaced by the following linear system of $3T$ equations
\begin{equation}
\begin{bmatrix} \boldsymbol{l_{M_{i_2}}^{M_{i_1}}}(t_1) \\ \vdots \\ \boldsymbol{l_{M_{i_2}}^{M_{i_1}}}(t_T) \end{bmatrix} = 
\begin{bmatrix} A\left(q_k(t_1)\right) \\ \vdots \\ A\left(q_k(t_T)\right) \end{bmatrix} \boldsymbol{b_{i_1,i_2}} = A(X)\boldsymbol{b_{i_1,i_2}} \text{.}
\label{eq:prism_lin}
\end{equation}%

We have the following Proposition.

\begin{prop} \label{prop:prism_lin}
Consider two markers $M_{i_1}$ and $M_{i_2}$, attached to consecutive links connected through a prismatic joint. Let $q_k$ be the joint signal influencing the relative motion between the two links. Then, given a set of observations $X$, the rotation matrix $R_{M_{i_2}}^{M_{i_1}}$ is constant and the linear system of equations in \eqref{eq:prism_lin} has solution satisfying the constraint in \eqref{eq:prism_constr}. 
\end{prop}

\subsection{Revolute joint}
Now assume that the joint connecting $L_{j_1}$ and $L_{j_2}$ is revolute. Expressing $\boldsymbol{l_{L_{j_2}}^{L_{j_1}}}$ as in \eqref{eq:joint_traslation}, we have that equation \eqref{eq:relative_position_consecutive markers} can be rewritten as 
\begin{align}\label{eq:Relative_Position_Consecutive_Markers}
\boldsymbol{l_{M_{i_2}}^{M_{i_1}}} 
=&\boldsymbol{l_{L_{j_1}}^{M_{i_1}}} + R_{L_{j_1}}^{M_{i_1}} \left(\begin{bmatrix} 0 \\ 0 \\d_{j_2} \end{bmatrix} + R_{L_{j_2}}^{L_{j_1}}(\theta_{j_2})\begin{bmatrix} a_{j_2} \\ 0 \\0 \end{bmatrix}\right) \notag \\ &+ R_{M_{i_2}}^{M_{i_1}}\left( -\boldsymbol{l_{L_{j_2}}^{M_{i_2}}} \right) \notag\\
=&\boldsymbol{l_{L_{j_1}}^{M_{i_1}}} + R_{L_{j_1}}^{M_{i_1}} \begin{bmatrix} 0 \\ 0 \\d_{j_2} \end{bmatrix} 
+ R_{M_{i_2}}^{M_{i_1}} R_{L_{j_2}}^{M_{i_2}} \begin{bmatrix} a_{j_2} \\ 0 \\0 \end{bmatrix} \notag \\ & 	+ R_{M_{i_2}}^{M_{i_1}}\left( -\boldsymbol{l_{L_{j_2}}^{M_{i_2}}} \right) \text{,}
\end{align}%
which is linear w.r.t. the vector of variables
\begin{equation*}
\boldsymbol{\bar{b}_{i_1,i_2}}= \begin{bmatrix}
\boldsymbol{l_{L_{j_1}}^{M_{i_1}}} + R_{L_{j_1}}^{M_{i_1}} \begin{bmatrix} 0 \\ 0 \\d_{j_2} \end{bmatrix} \\
R_{L_{j_2}}^{M_{i_1}} \begin{bmatrix} a_{j_2} \\ 0 \\0 \end{bmatrix} - \boldsymbol{l_{L_{j_2}}^{M_{i_2}}}\end{bmatrix} \text{,}
\end{equation*}%
since Equation \eqref{eq:Relative_Position_Consecutive_Markers} can be rearranged as
\begin{equation*}
\boldsymbol{l_{M_{i_2}}^{M_{i_1}}} = \begin{bmatrix}
I_3 & R_{M_{i_2}}^{M_{i_1}}
\end{bmatrix}  \begin{bmatrix}
\boldsymbol{l_{L_{j_1}}^{M_{i_1}}} + R_{L_{j_1}}^{M_{i_1}} \begin{bmatrix} 0 \\ 0 \\d_{j_2} \end{bmatrix} \\
R_{L_{j_2}}^{M_{i_1}} \begin{bmatrix} a_{j_2} \\ 0 \\0 \end{bmatrix} - \boldsymbol{l_{L_{j_2}}^{M_{i_2}}}\end{bmatrix} 
=  \bar{A}(D_p)\boldsymbol{\bar{b}_{i_1,i_2}} \text{.}
\end{equation*}%
When considering a set of observations $X$ with cardinality $T$ we obtain the following $3T$ liner equations 
\begin{equation}
\boldsymbol{l_{M_{i_2}}^{M_{i_1}}}(X) =
\begin{bmatrix}
\boldsymbol{l_{M_{i_2}}^{M_{i_1}}}(t_1) \\ \vdots \\ \boldsymbol{l_{M_{i_2}}^{M_{i_1}}}(t_T)
\end{bmatrix} = 
\begin{bmatrix}
\bar{A}(D_p^{t_1})\\ \vdots \\ \bar{A}(D_p^{t_T}) 
\end{bmatrix}\boldsymbol{\bar{b}_{i_1,i_2}} = \bar{A}(X)\boldsymbol{\bar{b}_{i_1,i_2}}
\label{eq:revolute_linear}
\end{equation}%
We have the following Proposition.
\begin{prop} \label{prop:revolute_lin}
	Consider two markers $M_{i_1}$ and $M_{i_2}$, attached to consecutive links connected through a revolute joint. Then, given a set of observations $X$, the linear system of equation in \eqref{eq:revolute_linear} has solution. 
\end{prop}

It is worth observing that the dependance of \eqref{eq:revolute_linear} on the revolute joint signal is not explicit since it is incorporated into the evolution of the matrix $R_{M_{i_2}}^{M_{i_1}}$. To directly consider the effects of varying the joint signal on the relative motion between $M_{i_1}$ and $M_{i_2}$, we analyze their relative orientation. Differently than in the prismatic case, $R_{M_{i_2}}^{M_{i_1}}$ is not constrained to be constant. Let $q_k$ be the actuation signal of the revolute joint between $L_{j_1}$ and $L_{j_2}$. 

Rewriting $R_{L_{j_2}}^{L_{j_1}}$ according to \eqref{eq:joint_orinentation}, we have that Equation \eqref{eq:relative_orientation_consecutive markers} can be rewritten as
\begin{align}
R_{M_{i_2}}^{M_{i_1}} &= R_{L_{j_1}}^{M_{i_1}} R_z(q_k+\theta_{j_2}^0) R_x(\alpha_{j_2}) R_{M_{i_2}}^{L_{j_2}} \notag\\
&=R_{L_{j_1}}^{M_{i_1}} R_z(q_k) R_{M_{i_2}}^{\bar{L}_{j_2}} \text{,}
\label{eq:RotationIdentity}
\end{align}%
where $R_{M_{i_2}}^{\bar{L}_{j_2}}=R_z(\theta_{j_2}^0)R_x(\alpha_{j_2}) R_{M_{i_2}}^{L_{j_2}}$. 
Analyzing \eqref{eq:RotationIdentity} element-wise, we can identify a system of nine equations where the unknowns are the elements of $R_{L_{j_1}}^{M_{i_1}}$ and $R_{M_{i_2}}^{\bar{L}_{j_2}}$, while the output is given by the elements of $R_{M_{i_2}}^{M_{i_1}}$. Now, let $Vec(\cdot)$ be the operator that maps a $N\times N$-dimensional matrix $M$ into the $N^2$- dimensional column vector obtained by stacking the columns of the matrix $M$ on top of one another, then we have that $Vec\left(R_{M_{i_2}}^{M_{i_1}}\right) = Vec \left(R_{L_{j_1}}^{M_{i_1}} R_z(q_k) R_{M_{i_2}}^{\bar{L}_{j_2}}\right)$.
In addition observe that the unknown variables must satisfy the following orthogonality constraints
\begin{align}
&Vec \left( R_{L_{j_1}}^{M_{i_1}} \left(R_{L_{j_1}}^{M_{i_1}}\right)^T \right) = Vec\left( I_3 \right) \notag \\
&Vec \left( R_{M_{i_2}}^{\bar{L}_{j_2}} \left(R_{M_{i_2}}^{\bar{L}_{j_2}}\right)^T \right) = Vec\left( I_3 \right) \label{eq:revolute_constr} \text{.}
\end{align}%
It is worth stressing that the aforementioned equations are non linear w.r.t the variables $Vec\left(R_{L_{j_1}}^{M_{i_1}} \right)$ and $Vec \left( R_{M_{i_2}}^{\bar{L}_{j_2}} \right)$. 

If, instead of considering a single observation we consider a set of observations $X$, the equations describing the relative orientations between $M_{i_1}$ and $M_{i_2}$ are
\begin{equation}
\begin{bmatrix} Vec\left(R_{M_{i_2}}^{M_{i_1}}(t_1)\right) \\ \vdots \\ Vec\left(R_{M_{i_2}}^{M_{i_1}}(t_T)\right) \end{bmatrix} = 
\begin{bmatrix}  Vec \left(R_{L_{j_1}}^{M_{i_1}} R_z\left(q_k(t_1)\right) R_{M_{i_2}}^{\bar{L}_{j_2}}\right) \\ \vdots \\  Vec \left(R_{L_{j_1}}^{M_{i_1}} R_z\left(q_k(t_T)\right) R_{M_{i_2}}^{\bar{L}_{j_2}}\right)\end{bmatrix} \label{eq:revolute_non_linear} \text{.}
\end{equation}%
We have the following proposition.
\begin{prop} \label{prop:revolute_non_lin}
	Consider two markers $M_{i_1}$ and $M_{i_2}$, attached to consecutive links connected through a revolute joint. Let $q_k$ be the actuation signal of the joint. Then, given a set of observations $X$, the system of non-linear equations in \eqref{eq:revolute_non_linear} admits a solution satisfying the non-linear orthogonality constraints in \eqref{eq:revolute_constr}. 
\end{prop}

\section{FULLY INFORMATIVE TRAJECTORIES} \label{sec:trj}

In the previous Section we have stated three propositions defining conditions that are verified when $M_{i_1}$ and $M_{i_2}$ are attached to consecutive links. In general the reverse relations are not true, since it is possible to exhibit sets of observations $X$ such that the conditions of the previous Propositions are satisfied though the markers are not attached to subsequent links; this fact is strictly related to the sequence of joint configurations which have generated the taken observations. Unfortunately, due to space constraints, we do not include in this paper examples of such \emph{false positive} observation sets. 

Instead, in this section we introduce a class of trajectories from which it is possible to properly select an observation set $X$ for which the conditions defined in Propositions 1 for the prismatic joint and in Propositions 2 and 3 for the revolute joint, are not only necessary but also sufficient to verify if two markers are attached to consecutive links. An observation set with this property is said to be \emph{fully informative}. A class of \emph{fully informative} observation sets can be formally defined as follows.


\begin{definition} \label{def:excitation_trj}
	Consider a collection of $n-1$ trajectories, where each trajectory is obtained moving only one joint, and keeping all the others stuck. For $k=1,\ldots, n-1$, without loss of generality, assume the $k$-th trajectory is obtained varying the joint signal $q_k$ and let $t_{1,k}$ and $t_{2,k}$ be two time instants such that $mod \left( q_k(t_{1,k}) \right) \neq mod \left( q_k(t_{2,k}) \right)$ and $q_w(t_{1,k}) = q_w(t_{2,k})$, $w \neq k$, where $mod(\cdot)$ is the $2\pi$ module operator. Then let us define the observation set $\bar{X}$ as the collection of the pairs
$$(D_p(t_{1,k}),\boldsymbol{q}(t_{1,k})), (D_p(t_{2,k}),\boldsymbol{q}(t_{2,k}))$$
for $k=1,\ldots n-1$.
\end{definition}

Observe that $\bar{X}$ has $2n-2$ observations, which for typical robot (i.e., $n=3,4,5,6$) represents a limited number of observations. It is possible to show that $\bar{X}$ is a fully informative set and in particular we have the following results.

\begin{prop} \label{prop:prism_optimal_q}
	Let $M_{i_1}$ and $M_{i_2}$ be two markers satisfying equations in Proposition \ref{prop:prism_lin} for a set of observations $\bar{X}$ defined in Definition \ref{def:excitation_trj} and joint signal $q_k$. Then the corresponding links $L_{j_1}$ and $L_{j_2}$ are subsequent in the kinematic chain and the joint between them is prismatic with input signal $q_k$. 
\end{prop}

\begin{prop} \label{prop:revolute_optim_q}
	Let $M_{i_1}$ and $M_{i_2}$ be two markers satisfying equations in Proposition \ref{prop:revolute_lin} and Proposition \ref{prop:revolute_non_lin} for a set of observations $\bar{X}$ defined in Definition \ref{def:excitation_trj} and joint signal $q_k$. Then the corresponding links $L_{j_1}$ and $L_{j_2}$ are subsequent in the kinematic chain and the joint between them is revolute with input signal $q_k$. 
\end{prop}

The proofs of the above Propositions are reported in Section \ref{sec:appendix}). 

A couple of remarks are now in order. 
\begin{remark}
As said there are examples of observation sets $X$ which are not fully informative and that might lead to false positive situations. However in the numerical Section we will show, by Monte Carlo simulations, that selecting these \emph{false positive} observation sets from generic trajectories seems to be a very unlikely event. \end{remark}
\begin{remark}
It is worth noting that the condition $mod \left( q_k(t_1) \right) \neq mod \left( q_k(t_2) \right)$ might not be verified with general input trajectories. A straightforward example happens when the actuation signal is periodic, with period $T_{q}$ and $t_2 = t_1 + T_q$. To avoid this situations we assume that the obtained trajectories are post processed, simply removing the redundant values.
\end{remark}

\section{PROPOSED APPROACH} \label{sec:proposed_approach}

\begin{figure}
	\centering
	\includegraphics[width=0.8\linewidth]{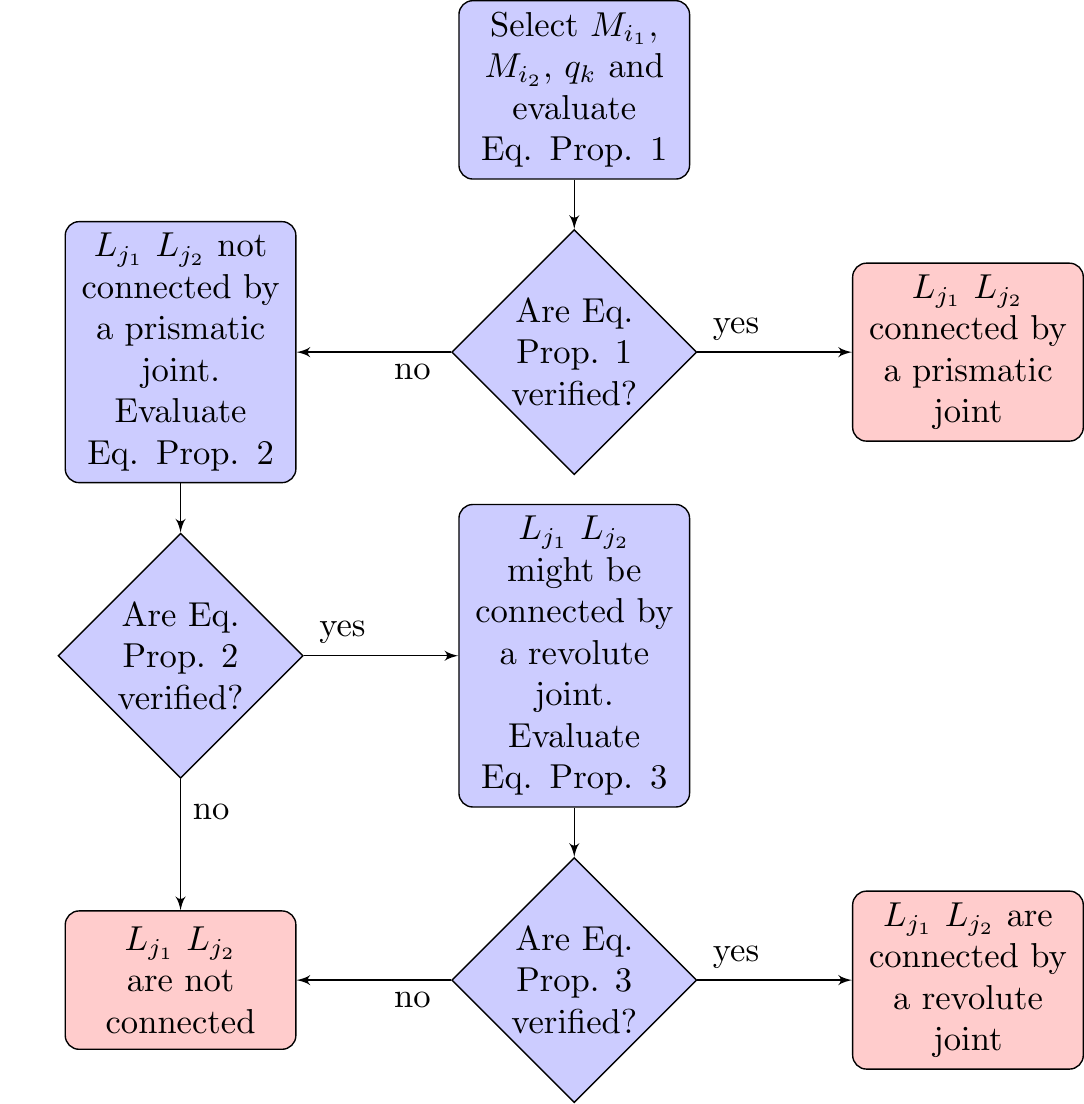}
	\caption{Flow chart of an iteration of the proposed algorithm. Ending conditions are highlighted in red.}
	\label{fig:FlowChart}
\end{figure}

In this Section we describe the algorithm we propose to deal with the robot kinematic structure identification problem. Our strategy consists in iterating over all the possible triplets, composed by a pair of markers and a joint signal, the  procedure described by the flow chart in Figure \ref{fig:FlowChart}.

Specifically, given a pair of markers and a joint signal, the algorithm first evaluates if the markers are connected through a prismatic joint and, in case this test is negative, it secondly evaluates if they are connected through a revolute joint. 

The first test consists in checking the feasibility of the linear equations defined in Proposition \ref{prop:prism_lin}. The second test, instead, is composed by two steps; the first step verifies if the linear equations of Proposition \ref{prop:revolute_lin} admit solution, and, in such case, also the second step is performed which consists in solving the system of non-linear equations of Proposition \ref{prop:revolute_non_lin}. Observe that, in this way, the last step is performed only when it is necessary, thus, minimizing its executions. This fact is particularly relevant from the computational point of view, since the non linear test is the most expensive.

\subsection{Empirical results for general trajectories} \label{sec:MC_exp}

In this section we investigate the effectiveness of the proposed approach for general trajectories, by running Monte Carlo simulations. In particular, simulating different robot kinematics (with $n=6$), we obtained a dataset composed of $128$ time-series, each one accounting for 50 observations. Among the different simulations we let vary several parameters, like the joint type order $S_Q$, the DH parameters and the markers positioning, namely $R_{M_i}^{L_j}$ and $\boldsymbol{l}_{M_i}^{L_j}$. As far as the input trajectories are concerned, we simulate a sinusoid for each joint signal, with amplitude and frequency randomly selected in each simulation.

For each time-series we have considered all the possible triplets, i.e.  all pairs of markers and joint signal, for a total number of $2400$ triplets, and we computed the systems of equations defined by Proposition \ref{prop:prism_lin}, \ref{prop:revolute_lin} and \ref{prop:revolute_non_lin}, to verify if the systems have solution.

\begin{figure}
	\centering
	\includegraphics[width=1\linewidth]{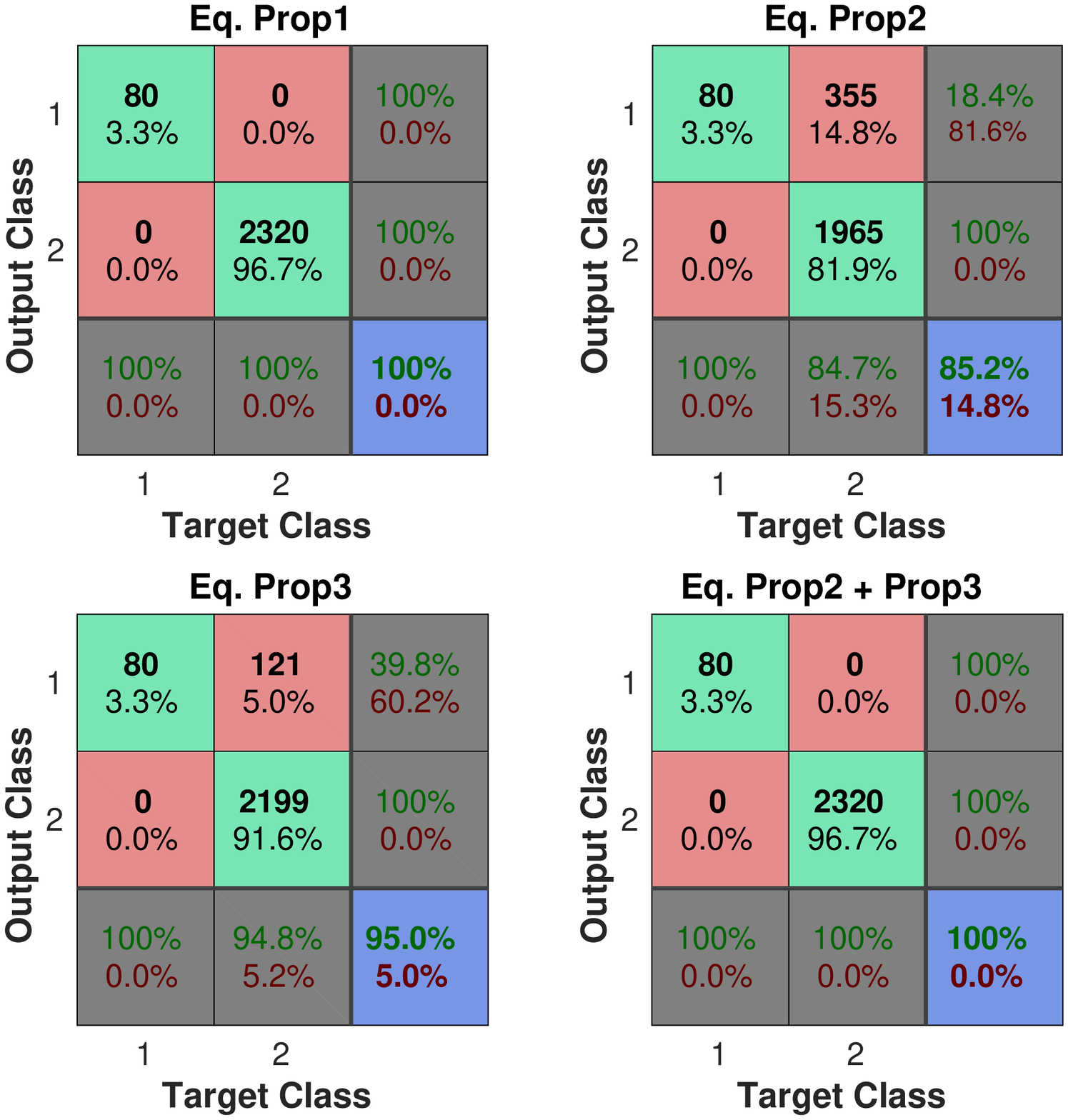}
	\caption{Confusion matrix of the Monte Carlo Simulation described in Section \ref{sec:MC_exp}. When considering Eq. Prop. \ref{prop:prism_lin}, a sample is assigned to Class $1$ if its markers and joint identify a prismatic transformation, while in the cases of Eq. Prop. \ref{prop:revolute_lin} and \ref{prop:revolute_non_lin} the elements in Class $1$ identify revolute transformations.
		}	
	\label{fig:conf_mat}
\end{figure}

Results are reported in Figure \ref{fig:conf_mat} in the form of confusion matrix. As usual the elements along the diagonal quantify the well classified triplets, while the elements outside the diagonal the ones which are misclassified. For example, when considering Proposition \ref{prop:prism_lin}, the $(1,1)$ entry quantifies the cases in which the system defined by Proposition \ref{prop:prism_lin} holds true and the elements of the considered triplets identify a prismatic transformation, while the $(2,2)$ entry the cases in which the system of equations is not verified and the relation between the elements of the triplet is not prismatic. The $(1,2)$ entry instead, represent the cases in which the system of equations holds even if the triple of elements are not connected, while the $(2,1)$ element represent the opposite situation. 

Results confirm that equations defined by Proposition \ref{prop:revolute_lin} and \ref{prop:revolute_non_lin}, identify only a necessary condition when considered alone, since a significant number of false positives occurred. Indeed it happens that the relative systems of equations have solutions even if the two markers are not connected. On the other hand, considering them together as done in Proposition \ref{prop:revolute_optim_q}, the number of false positives goes to zero, allowing a perfect classification.
Basically, empirical evidence highlights how the probability that the dataset collected contains only observations that leads to a false positives is close to zero.

\section{CONCLUSION} \label{sec:conclusion}
In this paper we introduced a novel and computational efficient algorithm able to learn the robot kinematic structure from visual observations. We prove the effectiveness of our approach in several simulated environments, testing different type of kinematics. As future work we plan to extend the proposed approach to the case of noisy measurements, and test the algorithm also in a real robot. In particular we expect to replace the two linear systems with two linear least squared problems, and the non linear system with a non linear least squared problem, as well as defining a threshold in the error, to discriminate among the cases in which the links are connected or not.

\section{APPENDIX} \label{sec:appendix}

To prove Proposition \ref{prop:prism_optimal_q} and \ref{prop:revolute_optim_q}, we need to introduce the relative transformations between two links $L_{j_1}$ and $L_{j_2}$ not subsequent in the kinematic chain. $R_{L_{j_2}}^{L_{j_1}}$, the relative orientation between $L_{j_1}$ and $L_{j_2}$ is obtained iterating equation \eqref{eq:joint_orinentation} along the kinematic chain, i.e. considering all transformations induced by the joints between $L_{j_1}$ and $L_{j_2}$:
\begin{small}
	\begin{equation}
	R_{L_{j_2}}^{L_{j_1}} = \prod_{m=j_1+1}^{m=j_2}\left( R_z(\theta_m)R_x(\alpha_m) \right)  \text{,}\label{eq:relative_orientation_chain} 
	\end{equation}%
\end{small}
where $\theta_m$ is a function of $q_{k_m}$ if the joint is revolute.

As regards the position the following general equation holds 
\begin{small}
	\begin{align}
	\boldsymbol{l_{L_{j_2}}^{L_{j_1}}} &= \sum_{m=j_1+1}^{m=j_2} \left( R_{L_{m-1}}^{L_{j_1}}\boldsymbol{l_{L_{m}}^{L_{m-1}}} \right) 
	\label{eq:relative_postion_chain} \text{.}
	\end{align}%
\end{small}

\subsection{Proof of Proposition \ref{prop:prism_optimal_q}}
Proposition \ref{prop:prism_optimal_q} states that, when considering the excitation trajectories in Definition \ref{def:excitation_trj} and the relative minimal set of observations $\bar{X}$, if $M_{i_1}$, $M_{i_2}$ and $q_k$ verify equations of Proposition \ref{prop:prism_lin}, then $L_{j_1}$ and $L_{j_2}$ are subsequent in the kinematic chain, and the joint between them is prismatic with $q_k$ as actuation signal.

We prove this proposition by contradiction, showing that, if the system of equations defined by Proposition \ref{prop:prism_lin} holds, then triple $M_{i_1}$, $M_{i_2}$ and $q_k$ describe a prismatic transformation.

First of all we exclude the case in which there is at least one revolute joint between $L_{j_1}$ and $L_{j_2}$. Without loss of generality let $j_3$ be the index of the link coming after the revolute joint. The relative orientation between the markers is $R_{M_{i_2}}^{M_{i_1}} = R_{L_{j_1}}^{M_{i_1}} R_{L_{j_3}}^{L_{j_1}} R_{L_{j_2}}^{L_{j_3}} R_{M_{i_2}}^{L_{j_2}}$, that, exploiting \eqref{eq:relative_orientation_chain}, becomes 
\begin{small}
	\begin{align}
	R_{L_{j_2}}^{L_{j_1}}=& R_{L_{j_1}}^{M_{i_1}}\prod_{m=j_1+1}^{m=j_3-1} \bigg( R_z(\theta_m)R_x(\alpha_m) \bigg)  R_z(\theta_{j_3}) R_x(\alpha_{j_3}) \notag \\
	& \prod_{\bar{m}=j_3+1}^{\bar{m}=j_2} \bigg( R_z(\theta_{\bar{m}})R_x(\alpha_{\bar{m}}) \bigg) R_{M_{i_2}}^{L_{j_2}} \label{eq:revolute_chain_dim}	 \\
	=& R_{L_{j_3-1}}^{M_{i_1}} R_z(\theta_{j_3}^0 + q_{k_{j_3}}) R_{M_{i_2}}^{\bar{L}_{j_3+1}} = R_{\bar{L}_{j_3-1}}^{M_{i_1}} R_z(q_{k_{j_3}}) R_{M_{i_2}}^{\bar{L}_{j_3+1}}\text{.} \notag
	\end{align}%
\end{small}

If Equations of Proposition \ref{prop:prism_lin} are verified, the relative orientation between $M_1$ and $M_2$ is constant, and consequently, considering two different time instants $t_1$ and $t_2$, $R_{M_{i_2}}^{M_{i_1}}(t_1)= R_{M_{i_2}}^{M_{i_1}}(t_2)$, that implies
\begin{small}
	\begin{align}
	&R_{\bar{L}_{j_3-1}}^{M_{i_1}} (t_1) R_z(q_{k_{j_3}}(t_1)) R_{M_{i_2}}^{\bar{L}_{j_3}}(t_1) = \notag \\
	&R_{\bar{L}_{j_3-1}}^{M_{i_1}} (t_2) R_z(q_{k_{j_3}}(t_2)) R_{M_{i_2}}^{\bar{L}_{j_3}}(t_2) \text{.} \label{eq:orientation_cond}
	\end{align}%
\end{small}%
However the last condition is not verified when the input trajectories are in $\bar{X}$. Indeed, when considering two input locations belonging to the subset of trajectories with $mod \left( q_{k_{j_3}}(t_1) \right) \neq mod \left( q_{k_{j_3}}(t_2) \right)$ and $q_w(t_1) = q_w(t_2)$ if $w \neq {k_{j_3}}$, $R_{\bar{L}_{j_3-1}}^{M_{i_1}} (t_1) = R_{\bar{L}_{j_3-1}}^{M_{i_1}} (t_2)$ and $R_{M_{i_2}}^{\bar{L}_{j_3}}(t_1) = R_{M_{i_2}}^{\bar{L}_{j_3}}(t_2)$, since they are function of the $q_w$ signals, that assume the same values in $t_1$ and $t_2$. This implies $R_z(q_{k_{j_3}}(t_1)) = R_z(q_{k_{j_3}}(t_2))$, and for the uniqueness if the Euler angles $mod \left( q_{k_{j_3}}(t_1) \right) = mod \left( q_{k_{j_3}}(t_2) \right)$, that is against Definition \ref{def:excitation_trj}.

The last observation proves that, when the joint signals are chosen in accordance with Definition \ref{def:excitation_trj}, $R_{M_{i_2}}^{M_{i_1}}$ can not be constant over $\bar{X}$ if there is at least one revolute joint between the links. Consequently, to conclude the proof, we consider a configuration in which there are one or more prismatic joints between $L_{j_1}$ and $L_{j_2}$. In these configurations the $M_{i_2}$ RF origin in the $M_{i_1}$ RF is%
\begin{small}
	\begin{align}
	\boldsymbol{l_{M_{i_2}}^{M_{i_1}}} =& \boldsymbol{l_{L_{j_1}}^{M_{i_1}}} + R_{L_{j_1}}^{M_{i_1}} \boldsymbol{l_{L_{j_2}}^{L_{j_1}}} + R_{M_{i_2}}^{M_{i_1}}\left( -\boldsymbol{l_{L_{j_2}}^{M_{i_2}}} \right)  \label{eq:prism_chain_non_lin} \\
	=&\sum_{m=j_1+1}^{m=j_2} \left( R_{L_{m-1}}^{M_{i_1}}  \left( \begin{bmatrix} 0\\0\\d_m \end{bmatrix} +  R_z(\theta_m)\begin{bmatrix}a_m\\0\\0\end{bmatrix} \right) \right) +  \notag\\
	& \boldsymbol{l_{L_{j_1}}^{M_{i_1}}} + R_{M_{i_2}}^{M_{i_1}}\left( -\boldsymbol{l_{L_{j_2}}^{M_{i_2}}} \right) +
	\sum_{m=j_1+1}^{m=j_2} \left( R_{L_{m-1}}^{M_{i_1}}  \begin{bmatrix} 0\\0\\q_{k_m} \end{bmatrix} \right) \text{,} \notag
	\end{align}%
\end{small}%
where the terms before the last sum are constant and equal to $\boldsymbol{l_{i_1,i_2}}$, while the ones in the last sum depends on the joint signals. As \eqref{eq:non_lin_prims_position}, the last equation defines a linear system with coefficient matrix $A^* = \left[ I_3 \,\,\,\, q_{k_{j_1}}I_3 \,\,\,\, \dots \,\,\,\, q_{k_{j_2}}I_3 \right]$ and vector of variables $\boldsymbol{b_{i_1,i_2}^*}= \left[ (\boldsymbol{l_{i_1,i_2}})^T \,\, Z(R_{L_{j_1}}^{M_{i_1}})^T \,\, \dots \,\, Z(R_{L_{j_2-1}}^{M_{i_1}})^T \right]^T$, where, as in \eqref{eq:non_lin_prims_position}, the $Z(R_{L_{m-1}}^{M_{m}})$ elements have unitary norm. Assuming equations of Proposition \ref{prop:prism_lin} are verified for the set of observations $\bar{X}$, it holds
\begin{small}
	\begin{equation}
	\boldsymbol{l_{M_{i_2}}^{M_{i_1}}}(\bar{X}) = A(\bar{X}) \boldsymbol{b_{i_1,i_2}} = A^*(\bar{X}) \boldsymbol{b_{i_1,i_2}^*} \text{.} \label{eq:prism_chain_lin}
	\end{equation}%
\end{small}
Anyway, if the inputs are selected in accordance with Definition \ref{def:excitation_trj}, $rank \left(A(\bar{X})\right)= 6$ and rank $rank \left(A^*(\bar{X})\right)= 3+3(j_2-j_1)$, and, given the constraints on the $\boldsymbol{b_{i_1,i_2}}$ and $\boldsymbol{b_{i_1,i_2}^*}$ elements, $rank \left(A(\bar{X}) \boldsymbol{b_{i_1,i_2}} \right) \leq rank \left(A^*(\bar{X}) \boldsymbol{b^*_{i_1,i_2}} \right)$. Moreover to obtain $rank \left(A(\bar{X}) \boldsymbol{b_{i_1,i_2}} \right) = rank \left(A^*(\bar{X}) \boldsymbol{b^*_{i_1,i_2}} \right)$ we need to consider only one prismatic joint and, since $span\left( \left[ q_{k_i}(t_1) \dots q_{k_i}(t_{n-1})\right] \right) = span\left( \left[ q_{k_j}(t_1) \dots q_{k_j}(t_{n-1})\right] \right)$ if and only of $k_i = k_j$, the equivalence in \eqref{eq:prism_chain_lin} holds if and only if the joint input signal is $q_k$, proving Proposition \ref{prop:prism_optimal_q}.

\subsection{Proof of Proposition \ref{prop:revolute_optim_q}}

As done for Proposition \ref{prop:prism_optimal_q}, we prove the proposition by contradiction. Assume that conditions defined by Proposition \ref{prop:revolute_lin} and Proposition \ref{prop:revolute_non_lin} are satisfied, and the inputs are selected accordingly to Definition \ref{def:excitation_trj}. Moreover suppose that $L_{j_1}$ and $L_{j_2}$ are not consecutive. Let $N_{rev}$ be the number of revolute joints present in the chain between the two links, and $k_{j_3} \dots k_{j_3+N_{rev}}$ their indexes. The relative orientation between $M_{i_2}$ and $M_{i_1}$ is $R_{M_{i_2}}^{M_{i_1}} = R_{L_{j_1}}^{M_{i_1}} R_{L{j_1}}^{L_{j_2}} R_{M_{i_2}}^{L_{j_2}}$, with $R_{L{j_1}}^{L_{j_2}}$ as in \eqref{eq:relative_orientation_chain}. As done in \eqref{eq:revolute_chain_dim}, the relative orientation between the markers can be rewritten highlighting the contributions of a particular revolute joint, for example the $j_3$ joint, obtaining $R_{M_{i_1}}^{M_{i_2}} = R_{L_{j_3}}^{M_{i_1}} R_z(q_{k_{j_3}}) R_{M_{i_2}}^{\bar{L}_{j_3}}$. At the same time, since conditions defined by Proposition \ref{prop:revolute_non_lin} are assumed to be true, it holds $R_{M_{i_1}}^{M_{i_2}} = R_{\hat{L}_{j_1}}^{\hat{M}_{i_1}} R_z(q_k) R_{\hat{M}_{i_2}}^{\hat{\bar{L}}_{j_2}}$, with $R_{\hat{L}_{j_1}}^{\hat{M}_{i_1}}$ and $R_{\hat{M}_{i_2}}^{\hat{\bar{L}}_{j_2}}$ corresponding to the solution of the non linear system. Then it follows 
\begin{small}
	\begin{equation*}
	R_{\hat{L}_{j_1}}^{\hat{M}_{i_1}} R_z(q_k) R_{\hat{M}_{i_2}}^{\hat{\bar{L}}_{j_2}} = R_{L_{j_3}}^{M_{i_1}} R_z(q_{k_{j_3}}) R_{M_{i_2}}^{\bar{L}_{j_3}} \text{,}
	\end{equation*}%
	and exploiting the standard properties of the rotation matrices and the Euler Angles, it holds $R_z(q_k) =  R_{\hat{L}_{\hat{M}_{i_1}}}^{\hat{L}_{j_1}} R_{L_{j_3}}^{M_{i_1}} R_z(q_{k_{j_3}}) R_{M_{i_2}}^{\bar{L}_{j_3}} R_{\hat{\bar{L}}_{j_2}}^{\hat{M}_{i_2}}$. Moreover, $R_{\hat{L}_{\hat{M}_{i_1}}}^{j_1} R_{L_{j_3}}^{M_{i_1}}$ and $R_{M_{i_2}}^{\bar{L}_{j_3}} R_{\hat{\bar{L}}_{j_2}}^{\hat{M}_{i_2}}$ can be expressed with respect to their Euler Angles, for example adopting the $zxz$ convention, obtaining $R_z(q_k) =  R_z(\alpha_1)R_x(\beta_1)R_z(\gamma_1) R_z(q_{k_{j_3}}) R_z(\alpha_2)R_x(\beta_2)R_z(\gamma_2)$ and then
	\begin{align*}
	R_z(q_k-\alpha_1-\gamma_2) =  R_x(\beta_1) R_z(q_{k_{j_3}}+\gamma_1+\alpha_2) R_x(\beta_2) \text{.}
	\end{align*}%
\end{small}%
When neglecting the degenerate configurations, for the Euler Angles uniqueness the last equation is true if and only if $\beta_1$ and $\beta_2$ are null, and $q_k-\alpha_1-\gamma_2 = q_{k_{j_3}}+\gamma_1+\alpha_2$, namely $q_k-q_{k_{j_3}} = \alpha_1+\gamma_2+\gamma_1+\alpha_2$. Now consider two time instants $t_1$ and $t_2$ in $\bar{X}$ satisfying $mod \left( q_{k_{j_3}}(t_1) \right) \neq mod \left( q_{k_{j_3}}(t_2) \right) $ and $q_w(t_1) = q_w(t_2)$ if $w \neq k_{j_3}$. It holds  $R_{L_{j_3}}^{M_{i_1}}(t_1) = R_{L_{j_3}}^{M_{i_1}}(t_2)$ and $R_{M_{i_2}}^{\bar{L}_{j_3}}(t_1) = R_{M_{i_2}}^{\bar{L}_{j_3}}(t_2)$, implying also that $\alpha_1$, $\beta_1$, $\gamma_1$, $\alpha_2$, $\beta_2$ and $\gamma_2$ are constant in $t_1$ and $t_2$. Then
\begin{small}
	\begin{equation*}
	q_k(t_1)-q_{k_{j_3}}(t_1) = q_k(t_2)-q_{k_{j_3}}(t_2) \text{.}
	\end{equation*}%
\end{small}%
Under the excitation assumptions in Definition \ref{def:excitation_trj}, the last equation holds if and only if $k=k_{j_3}$. Similar considerations can be done considering the other revolute joints and opportune time instants, resulting in a set of conditions that are unfeasible when $N_{rev}>1$. This observations prove two important facts which are verified if the system of equations defined by Proposition \ref{prop:revolute_non_lin} holds for $\bar{X}$. The first fact is that only one revolute transformation is allowed, and the second is that the input of the revolute joint must be $q_k$.

To conclude the analysis we need to exclude the case in which there is a revolute joint with input signal $q_k$ and one or more prismatic joints between $L_{j_1}$ and $L_{j_2}$. This case is excluded by equations of Proposition \ref{prop:revolute_lin}. Let $j_3$ be the revolute joint index. Consider the $\boldsymbol{l_{M_{i_1}}^{M_{i_2}}}$ expression in \eqref{eq:relative_position_consecutive markers}, and expand $\boldsymbol{l_{L_{j_2}}^{L_{j_1}}}$ in order to highlight the prismatic transformations occurring before or after the revolute joint $\boldsymbol{l_{M_{i_1}}^{M_{i_2}}}$. Then it follows
\begin{small}
	\begin{align}
	\boldsymbol{l_{M_{i_1}}^{M_{i_2}}} =& \boldsymbol{l_{L_{j_1}}^{M_{i_1}}} + R_{M_{i_2}}^{M_{i_1}}(-\boldsymbol{l_{L_{j_2}}^{M_{i_2}}}) + \sum_{m=j_1+1}^{m=j_3-1}R_{L_{j_m}}^{M_{i_1}} \boldsymbol{l_{L_{m}}^{L_{m-1}}} \notag\\
	&+ R_{M_{i_2}}^{M_{i_1}}\sum_{\bar{m}=j_3+1}^{\bar{m}=j_2}R_{L_{j_{\bar{m}-1}}}^{M_{i_2}} \boldsymbol{l_{L_{\bar{m}}}^{L_{\bar{m}-1}}} \text{,} \label{eq:rev_contradiction}
	\end{align}%
\end{small}%
where the $R_{L_{j_m}}^{M_{i_1}}$ and $R_{L_{j_{\bar{m}-1}}}^{M_{i_2}}$ matrices in the sums are constant since they refer to prismatic transformations, while the $\boldsymbol{l_{L_{m}}^{L_{m-1}}}$ and $\boldsymbol{l_{L_{\bar{m}}}^{L_{\bar{m}-1}}}$ expressions are defined in \eqref{eq:joint_traslation} with $d_j$ dependent on the joint signals. As in equation \eqref{eq:prism_chain_non_lin}, the contributions due to the first sum and the first term define a linear system with coefficient matrix $A^* = \left[ I_3 \,\,\,\, q_{k_{j_1}}I_3 \,\,\,\, \dots \,\,\,\, q_{k_{j_2}}I_3 \right]$, and vector of variables $\boldsymbol{b_{i_1,j_3}^*}= \left[ (\boldsymbol{l_{i_1,j_3}})^T \,\, Z(R_{L_{j_1}}^{M_{i_1}})^T \,\, \dots \,\, Z(R_{L_{j_3-1}}^{M_{i_1}})^T \right]^T$, with the $Z(R_{L_{m-1}}^{M_{m}})$ elements constrained to have unitary norm. On the other hand, when Proposition \ref{prop:revolute_lin} holds, $\boldsymbol{l_{M_{i_1}}^{M_{i_2}}} \in span\left(\bar{A}(\bar{X})\right)$, and considering the constraints on the $\boldsymbol{b_{i_1,j_3}^*}$ norm and that $span\left(A^*(\bar{X})\right) \notin span\left(\bar{A}(\bar{X})\right)$, it follows that it is impossible that there are prismatic transformations between $L_{j_1}$ and the link after the revolute joint, i.e. $j_3=j_1+1$. To conclude the proof we need to exclude the possibility that there are prismatic transformations between the revolute joint and $L_{j_2}$. Observe that the second sum contributions in \eqref{eq:rev_contradiction} are proportional to $R_{M_{i_2}}^{M_{i_1}}$ with time-variant coefficient, due to the joint signals variability. Anyway this is in contradiction with the conditions in Proposition \ref{prop:revolute_lin} and the proof is concluded. Indeed, since $\bar{A}(\bar{X})$ is a full rank matrix, if \eqref{eq:revolute_linear} holds, its solution is unique, but this is not in accordance with the coefficient variability in \eqref{eq:rev_contradiction}.

\bibliographystyle{IEEEtran}
\bibliography{references}

\begin{thebibliography}{10}
\providecommand{\url}[1]{#1}
\csname url@samestyle\endcsname
\providecommand{\newblock}{\relax}
\providecommand{\bibinfo}[2]{#2}
\providecommand{\BIBentrySTDinterwordspacing}{\spaceskip=0pt\relax}
\providecommand{\BIBentryALTinterwordstretchfactor}{4}
\providecommand{\BIBentryALTinterwordspacing}{\spaceskip=\fontdimen2\font plus
\BIBentryALTinterwordstretchfactor\fontdimen3\font minus
  \fontdimen4\font\relax}
\providecommand{\BIBforeignlanguage}[2]{{%
\expandafter\ifx\csname l@#1\endcsname\relax
\typeout{** WARNING: IEEEtran.bst: No hyphenation pattern has been}%
\typeout{** loaded for the language `#1'. Using the pattern for}%
\typeout{** the default language instead.}%
\else
\language=\csname l@#1\endcsname
\fi
#2}}
\providecommand{\BIBdecl}{\relax}
\BIBdecl

\bibitem{Mod_rob_1}
M.~Yim, D.~G. Duff, and K.~D. Roufas, ``Polybot: a modular reconfigurable
  robot,'' in \emph{Proceedings 2000 ICRA. Millennium Conference. IEEE
  International Conference on Robotics and Automation. Symposia Proceedings
  (Cat. No.00CH37065)}, vol.~1, April 2000, pp. 514--520 vol.1.

\bibitem{Mod_rob_2}
M.~Yim, W.~Shen, B.~Salemi, D.~Rus, M.~Moll, H.~Lipson, E.~Klavins, and G.~S.
  Chirikjian, ``Modular self-reconfigurable robot systems [grand challenges of
  robotics],'' \emph{IEEE Robotics Automation Magazine}, vol.~14, no.~1, pp.
  43--52, March 2007.

\bibitem{Mod_rob_3}
L.~Brodbeck and F.~Iida, ``Enhanced robotic body extension with modular
  units,'' in \emph{2012 IEEE/RSJ International Conference on Intelligent
  Robots and Systems}, Oct 2012, pp. 1428--1433.

\bibitem{latombe}
J.-C. Latombe, \emph{Robot Motion Planning}.\hskip 1em plus 0.5em minus
  0.4em\relax Norwell, MA, USA: Kluwer Academic Publishers, 1991.

\bibitem{siciliano}
B.~Siciliano, L.~Sciavicco, L.~Villani, and G.~Oriolo, \emph{Robotics,
  Modelling, Planning and Control}, 2009.

\bibitem{kin_calibration_PSW}
G.~Gao, W.~Wang, K.~Lin, and Z.~Chen, ``Kinematic calibration for articulated
  arm coordinate measuring machines base on particle swarm optimization,'' in
  \emph{2009 Second International Conference on Intelligent Computation
  Technology and Automation}, vol.~1, Oct 2009, pp. 189--192.

\bibitem{kin_point_cloud}
T.~Zhou and B.~E. Shi, ``Simultaneous learning of the structure and kinematic
  model of an articulated body from point clouds,'' in \emph{2016 International
  Joint Conference on Neural Networks (IJCNN)}, July 2016, pp. 5248--5255.

\bibitem{kin_end_effector}
M.~Hersch, E.~L.~Sauser, and A.~Billard, ``Online learning of the body
  schema.'' vol.~5, pp. 161--181, 06 2008.

\bibitem{Sturm_object}
T.~Rühr, J.~Sturm, D.~Pangercic, M.~Beetz, and D.~Cremers, ``A generalized
  framework for opening doors and drawers in kitchen environments,'' in
  \emph{2012 IEEE International Conference on Robotics and Automation}, May
  2012, pp. 3852--3858.

\bibitem{Sturm_robot}
J.~Sturm, C.~Plagemann, and W.~Burgard, ``Unsupervised body scheme learning
  through self-perception,'' in \emph{2008 IEEE International Conference on
  Robotics and Automation}, May 2008, pp. 3328--3333.

\bibitem{fiducial_markers}
M.~Fiala, ``Designing highly reliable fiducial markers,'' \emph{IEEE
  Transactions on Pattern Analysis and Machine Intelligence}, vol.~32, no.~7,
  pp. 1317--1324, July 2010.

\bibitem{apriltag1}
E.~Olson, ``Apriltag: A robust and flexible visual fiducial system,'' in
  \emph{2011 IEEE International Conference on Robotics and Automation}, May
  2011, pp. 3400--3407.

\bibitem{apriltag2}
J.~Wang and E.~Olson, ``Apriltag 2: Efficient and robust fiducial detection,''
  in \emph{2016 IEEE/RSJ International Conference on Intelligent Robots and
  Systems (IROS)}, Oct 2016, pp. 4193--4198.

\bibitem{marker_from_feature_1}
S.~Kim and K.~Yoon, ``Point density-invariant 3d object detection and pose
  estimation,'' in \emph{2017 IEEE International Conference on Image Processing
  (ICIP)}, Sept 2017, pp. 2647--2651.

\bibitem{marker_from_feature_2}
W.~Yun, J.~Lee, J.~Lee, and J.~Kim, ``Object recognition and pose estimation
  for modular manipulation system: Overview and initial results,'' in
  \emph{2017 14th International Conference on Ubiquitous Robots and Ambient
  Intelligence (URAI)}, June 2017, pp. 198--201.

\bibitem{marker_from_feature_3}
H.~Kim, J.~Y. Lee, J.~H. Kim, J.~B. Kim, and W.~Y. Han, ``Object recognition
  and pose estimation using klt,'' in \emph{2012 12th International Conference
  on Control, Automation and Systems}, Oct 2012, pp. 214--217.

\end{thebibliography}

\end{document}